\documentclass[cameraready]{Interspeech}

\usepackage{amsmath,graphicx,hyperref}
\usepackage[normalem]{ulem}
\usepackage{booktabs}
\usepackage{tabularx}
\usepackage{cite}

\title{Audio-Based Understanding of Audiobook Narration Appeal}

\author[affiliation={1,2}, orcid=0009-0007-5958-277X]{Shahar}{Elisha}
\author[affiliation={1}, orcid=0000-0002-8750-8346]{Mariano}{Beguerisse-D\'iaz}
\author[affiliation={2}, orcid=0000-0002-6820-6764]{Emmanouil}{Benetos}

\address{
    $^1$Spotify, $^2$Queen Mary University of London
}

\email{shahar@spotify.com}


\usepackage{comment}

\begin{document}

\maketitle

\begin{abstract}
    Narration is central to the audiobook listening experience, shaping how listeners engage with and understand the content. This work explores how narration qualities shape an audiobook’s appeal, noting that their effects can vary by genre, title, and audience. We extract vocal and acoustic features (e.g., tone, pace, loudness) from LibriVox using pre-trained audio models and analyse their relationship with consumption data (specifically, view-rate) and their interplay with genre and title. Despite limited consumption data, we find that acoustic information alone has a robust association with appeal, even after accounting for title effects. We further validate these findings using more nuanced proprietary engagement metrics. To our knowledge, this is the first systematic computational study linking narration qualities, genre, title, and audiobook consumption, highlighting the potential of data-driven insights to improve audiobook personalisation and narrator casting.
\end{abstract}

\vspace{-4pt}
\section{Introduction}
\label{sec:intro}

Narration style and acoustic presentation are important components of audiobooks; they have the power to either elevate or undermine a listener's experience, understanding, and engagement with the story~\cite{audiobooks_role_of_narrator}. While the narration alone may not be the determining factor in audiobook selection amongst users, it has a significant impact on whether a user completes listening to the book in full~\cite{Dakic1396778}. A narrator offers an interpretation of the story by varying their voice across pitch, volume, timbre, and tempo, portraying characteristics such as gender, age, accent, emotional states and attitudes~\cite{experiencing_kosch_2024}. This can range from giving each character a distinct voice, to maintaining a consistent tone throughout. Other acoustic qualities such as the recording quality and the use of music and sound effects, can give variety and texture to a narration. Different genres may call for different narration styles; for example, an immersive and expressive style may be better suited to fiction than to non-fiction or academic texts. 

Audiobooks are widely consumed, with platforms such as Spotify, Audible, LibriVox, and Libby offering extensive libraries. As catalogues expand, these platforms face unique challenges: distinguishing between multiple recordings of the same title, understanding the narrations that appeal to different users or suit different genres, and informing casting decisions. User-item interaction data is also sparser than in domains such as music - for example, audiobooks are rarely replayed, and users consume far fewer than songs. Improving classification, search, recommendation and personalisation models requires a sophisticated understanding of audiobooks, not just on its content (e.g., text, topic, genre, author), but also its narration style, and their interplay. Even small improvements in modelling appeal can yield substantial gains in large-scale recommendation systems, where marginal ranking improvements can translate into increased user engagement and retention~\cite{spotify_recs, netflix_recs}.

Here, we investigate the relationship between an audiobook's narration and acoustic features, genres, title, and consumption to understand how narration influences appeal. We analyse vocal and acoustic properties of narrations on a large, publicly available dataset of audiobooks, exploring nuances related to genres, differences within groups of the same texts, and consumption statistics. To our knowledge, this is the first computational study linking acoustic narration features to large-scale, real-world consumption data across genres and multiple recordings of the same title (see Sec. \ref{sec:related_works} for comparison to prior work). Specifically, our contributions include:
\begin{itemize}
    \item \vspace{-2pt} \textbf{Statistical analysis of audiobook consumption} to assess the influence of interpretable acoustic features on a large real-world dataset.
    \item \vspace{-3pt}\textbf{Genre-specific modelling} to reveal how and which acoustic features influence appeal across genres.
    \item \vspace{-3pt} \textbf{Intra-title comparison framework} for different narrations of the same book (i.e. book-groups) to control for content.
    \item \vspace{-3pt} \textbf{Classification and ranking models} to evaluate the predictive power of acoustic features for appeal.
\end{itemize}

\vspace{-1pt} We find a statistically significant relationship between the acoustic features of a narration and an audiobook's appeal. This result is striking for two reasons: the narration style is independent of the book content, and the dataset has substantial limitations, which include variable recording quality of volunteer narrations and coarse, limited consumption data. To assess the robustness of our findings, we conduct follow-up experiments replacing the LibriVox consumption proxy with more granular engagement metrics from a proprietary Spotify dataset. These results are an important step towards understanding audiobook narrations and their appeal, offering insights to improve narrator casting and recommendation systems by matching the right voice to the right book, and the right book to the right listener. 

\vspace{-4pt}
\section{Related Works}
\label{sec:related_works}

\subsection{Computational Paralinguistics and Voice Perception}
\vspace{-4pt}
Human voices carry paralinguistic information from which a listener perceives qualities about the speaker’s identity and intention~\cite{neurocomputational_krger_2018}. Researchers have developed computational models for paralinguistic tasks such as perceived gender and age classification, health predictors, emotions, social standings, vocal qualities, speaking styles, and even vocal attractiveness (e.g.,~\cite{affective_weningerd_2022, ser_generalisation_goron_2024, multidimensional_obuchi_2021, scripted_spontaneous_elisha_2024, JALALINAJAFABADI2021103018, spectral_flux_gender, charisma, depression}). The classic approach is to train machine learning models on predicted acoustic features from the audio (e.g., pitch, loudness, or spectral)~\cite{schuller2013_computational_paralinguistics}. Recent works employ sophisticated techniques, such as transformer models, for improved performance at the cost of interpretability~\cite{scripted_spontaneous_elisha_2024, ser_generalisation_goron_2024, ethical_awareness_batliner_2023}.

\vspace{-6pt}
\subsection{Narration Styles}

\vspace{-4pt}
While plenty of audiobook research uses Text to Speech (TTS) and voice conversion (VC) models to deliver expressive speech fit for storytelling (e.g.,~\cite{audiobook_tts, sini-etal-2022-investigating}), studies indicate that listeners prefer human narrations over synthetic ones~\cite{synthetic_rodero_2021}, although some are interested in the ability to customise synthetic narrations~\cite{evaluating_szkely_2012}. User surveys and qualitative frameworks establish the importance of narrator style to the audiobook experience~\cite{audiobooks_role_of_narrator, Dakic1396778, experiencing_kosch_2024}, but stop short of identifying which acoustic features drive appeal.
We build on computational studies exploring acoustic and vocal features of narration at different levels: utterance, dialogue, and narrator. Prosody and voice quality are salient discriminators between storytelling discourse modes (e.g., narrative, descriptive, and dialogue)~\cite{role_montao_2016}, a finding that holds cross-lingually~\cite{role_montao_2017}, though both approaches rely on small datasets. At the dialogue level, female characters are narrated with higher pitch and lower volume relative to males~\cite{prosody_pethe_2023}. Other studies cluster narrators by speaking style using glottal source parameters~\cite{clustering_szkely_2011} or Convolutional Neural Network (CNN) embeddings~\cite{embretsen2019representing}. Most relevant to our work, Lange et al.~\cite{narrative_lange_2020} link acoustic features to user-reported absorption and liking, finding higher articulation rates predict both, though feature correlations leave the specific drivers unclear. Critically, none of these studies examine narration appeal at scale across genres and alternative narrations of the same title.

\vspace{-4pt}
\section{Experimental Setup}
\label{sec:experimental_setup}

\vspace{-2pt}
\subsection{LibriVox catalogue}
\label{sec:librivox}

\vspace{-4pt}
LibriVox~\cite{librivox} is a catalogue of public domain audiobooks, read and recorded by volunteers, with multiple titles and genres. The metadata (e.g., title, author, narrator, genres, text-source) and audio files are available to download freely. The Internet Archive keeps track of the number of page views, favourites, and reviews of each recording~\cite{internetarchive_librivox}. For simplicity, we focus on single-narrator, English audiobooks. Our data contains 8,854 audiobooks read by 1,206 different narrators, across 65 genres (e.g., History, Romance, Comedy). We group audiobooks into \textit{book-groups} if they share the same link to the original text source (i.e., different recordings of the same text). The dataset can be found in the supplementary codebase\footnote{\url{https://github.com/spotify-research/audiobook-narrations-interspeech}}.
We segment recordings (typically a chapter) into 30s intervals; we sample up to 20 segments per recording (the first, last, and 18 random segments in the middle), giving a maximum of 10 minutes of audio per recording (shorter recordings are included in full). This strategy balances representativeness with computational efficiency.
In the absence of publicly available information about listener satisfaction, downloads, hours listened, or completion rates, we use the number of views as a proxy for appeal; the number of favourites and reviews are too small to be useful in this analysis. We divide the views by the number of days since publication on LibriVox (i.e. \textit{view-rate}) to account for time-on-platform bias. Note that view-rate is limited as an appeal metric (e.g., no distinction between completed listens, brief listens, or multiple listens by a user), which introduces additional noise to our analysis. Although an initial page view is recorded before any listening occurs, we assume that the view count reflects appeal through repeated visits, as audiobooks are seldom finished in a single sitting. This proxy is therefore biased against shorter recordings, which may require fewer visits to complete and thus accumulate lower view-rates independent of appeal. Nevertheless, it is the only large-scale proxy available, and we use it to ensure transparency and reproducibility. We further extend our analysis using more nuanced engagement metrics from proprietary Spotify data to evaluate the robustness of our findings under a less coarse proxy (Sec.~\ref{sec:proprietary_experiment}).

\vspace{-6pt}
\subsection{Feature extraction and aggregation}

\vspace{-4pt}
We extract acoustic and prosodic features for every 30s sample using well-known, pre-trained audio models with default parameters (see Table~\ref{tab:features}), and we concatenate the features for all segments and chapters per audiobook along the time axis and calculate summary statistics:
\begin{itemize}
    \item \textbf{\textit{eGeMAPSv02}}~\cite{opensmile_gemaps}: 25 low-level acoustic descriptors of \textit{frequency}, \textit{energy}, \textit{spectral}, and \textit{tempo} using the openSMILE tool~\cite{opensmile}. We replicate openSMILE's aggregation logic to output 84 functional features (see codebase\footnotemark[1]).
    \item \textbf{\textit{YAMNet}}~\cite{yamnet} scores: 521 \textit{audio events} classification scores, such as human vocals, musical instruments, animal and environmental sounds.
     Using the AudioSet ontology~\cite{audioset}, we group audio classes into \textit{sound effects}, \textit{music}, \textit{recording quality}, \textit{nonverbal vocalisations}, and \textit{speech}. We keep the 13 individual \textit{speech} subclasses, and aggregate the remaining 4 grouped audio classes by taking the maximum subclass activation, giving a total 17 audio events. We calculate the mean and standard deviation of the scores across time for a final audio events set of 34 features. 
    \item \textbf{\textit{whisper-tiny}}~\cite{whisper} transcripts with timestamps. We extract syllables~\cite{syllables_package} and compute word/syllable counts and rates. We compute the mean, std, min and max of the word and syllable rates, and the total duration, word and syllable counts, giving 11 features.
\end{itemize}
Finally, we concatenate the features into a 129-dimensional audiobook vector (see Table 1 in Supplementary Materials (SM)\footnotemark[1]).

\begin{table}[t]
    \centering
    \footnotesize
    \setlength{\tabcolsep}{4pt}      
    \renewcommand{\arraystretch}{1} 
    \caption{Acoustic and vocal feature classes, see SM:Table 1 for all 129 summary statistics.}
    \begin{tabular}{@{} p{0.21\linewidth} p{0.75\linewidth} @{}}
        \toprule
        \textbf{Descriptors} & \textbf{Features} \\
        \midrule
        Frequency &
        F0, jitter, F1--3, F1--3 bandwidth \\
        Energy / amplitude &
        Loudness, shimmer, harmonics-to-noise ratio (HNR), equivalent sound level, loudness peaks per second \\
        Spectral &
        Alpha ratio, Hammarberg index, spectral slope, F1--3 relative energy, harmonic differences (H1--H2, H1--H3), MFCCs 1--4, spectral flux \\
        Tempo &
        Word rate, syllable rate, duration \\
        Audio events &
        Music, sound effects, recording quality, non-verbal vocalisations, speech (e.g.\ conversation, narration/monologue, screaming, whispering) \\
        \bottomrule
    \end{tabular}
    \label{tab:features}
\end{table}

\vspace{-6pt}
\subsection{Statistical Modelling}

\vspace{-4pt}
We investigate how acoustic and vocal features relate to appeal (view-rate), including genre-specific and intra-title nuances. We report model fit statistics and statistically significant coefficients. We apply the Benjamini-Hochberg (BH) method to correct for false correlations ($p < 0.05$)~\cite{benjamini-hochberg}. \newline
{\bf \noindent Feature pre-processing}:
To address multicollinearity in our feature set, we perform Variance Inflation Factor (VIF) pruning~\cite{montgomery2021introduction}, by iteratively removing the feature with the highest VIF score (i.e. highly correlated with other features) until all features have a factor below 5. We standardise the resulting 70 features prior to model fitting. \newline
{\bf \noindent Global modelling of consumption}:
We fit a Generalised Linear Model (GLM)~\cite{montgomery2021introduction} using the pruned and standardised acoustic features to model audiobook view-rate. The GLM allows us to assess the significance of individual features and identify which acoustic properties are most predictive of audiobook popularity. To account for the long tail in the view-rate distribution, we log-transform the view-rate and assume a Gaussian error distribution. \newline
{\bf \noindent Genre-specific}:
To investigate genre-specific effects, we train a GLM per genre (65 genres). Within each genre, we re-standardise the feature set to control for genre-specific distributions. We set the initial parameters to those learned by the global GLM, and drop near-constant features ($std\leq1\mathrm{e}{-12}$). We then compare BH-corrected significant coefficients across genres to examine whether acoustic correlates of appeal vary by genre. \newline
{\bf \noindent Intra-title}:
To account for title-specific effects on appeal, we fit a Linear Mixed-Effects (LME) model~\cite{montgomery2021introduction}. We model a random intercept per \textit{book-group}, while the acoustic features serve as the fixed effects. This allows us to disentangle title-driven variability from acoustic influences on view-rate. We transform our view-rate into log scale to align with the GLMs. We compare the model (retrained using maximum likelihood) to the global GLM using Akaike Information Criterion (AIC)~\cite{aic}, a relative measure that helps identify the better-fitting model.

\vspace{-6pt}
\subsection{Predictive Modelling}
\label{sec:pred_modelling}

\vspace{-4pt}
We also want to understand the predictive power of acoustic features on classifying and ranking appeal. Because the view-rate metric provides limited nuance (see Sec.~\ref{sec:librivox}), we simplify the problem space by binning view-rates into quartiles, thus converting the task from regression to classification. We use all 129 features as input and standardise during training. We run bootstrapping to estimate the 95\% confidence intervals (CIs) and a Kolmogorov-Smirnov test to confirm that the model predictions are better than random. We run 1000 random realisations, and 1000 bootstrapping iterations. \newline
{\bf \noindent Classification}:
Given the small and largely interpretable feature set, we train four shallow classifiers: Logistic Regression (LR), Support Vector Machine (SVM), XGBoost, and Multi-Layer Perceptron (MLP). For comparison, we train classification models using genres as input (multi-hot encoding), and a combination of acoustic features and genres.  We train and evaluate models with 5-fold cross-validation, grouped by narrator and balanced by view-rate quartile. We evaluate overall accuracy and assess performance across view-rate quartiles. \newline
{\bf \noindent Ranking}:
To capture appeal between different narrations of the same title (i.e. \textit{book-groups}), we train ranking models using acoustic features to predict the within-title rankings where an audiobook with a higher view-rate ranks higher within its group. We filter the data to books in groups of size $\geq 2$, resulting in 305 groups and 736 audiobooks. Most groups are of size 2, and the maximum group size is 13. We train and evaluate models with 5-fold cross-validation, grouped by book-groups. We train three boosted tree models: XGBRanker~\cite{xgboost_docs} with a rank:ndcg objective (based on LambdaMART~\cite{lambdamart}), with a rank:pairwise objective (uses RankNet loss~\cite{ranknet}), and LGBMRanker~\cite{lightgbm_docs} (based on LambdaRank)~\cite{burges2010from}. While the default training objectives use Normalised Discounted Cumulative Gain (NDCG), we report performance using average Kendall's rank correlation coefficients (Kendall's $\tau$) between predicted and true book-group rankings, as it directly measures order correlation and avoids inflated NDCG scores due to small, fully relevant groups (the average NDCG on random rankings for this dataset is 0.92). Although differentiable adaptations of Kendall’s $\tau$ exist~\cite{diffkendall}, we retain the standard training objectives as a starting point since they are well-established, and focus on Kendall’s $\tau$ purely for evaluation. We compare performance to two pointwise models using the LR model implemented for classification: 1) trained and evaluated on the 736 books used for the ranking models, and 2) trained and evaluated on the full dataset of 8,854 books, with different cross-validation splits.

\vspace{-4pt}
\section{Results}
\label{sec:results}

\vspace{-4pt}
\subsection{Statistical Modelling Results}

\vspace{-4pt}
{\bf \noindent Global modelling of consumption}:
The GLM attains a pseudo-$R^{2}$ of 0.09, indicating that narration-related properties explain a measurable portion of variation in appeal despite the coarse proxy (see Sec.~\ref{sec:librivox}) and omission of title, genre, and promotional factors. In a large and noisy real-world dataset, explaining nearly 10\% of the variability using audio features alone suggests that narration characteristics have a consistent and non-trivial association with listener engagement. We find 31 acoustic features show a significant effect on view-rate (see SM:Fig. 1), though no single feature dominates and effect sizes are mostly small ($\lvert\beta\rvert \leq 0.13$), suggesting that appeal is shaped by a combination of acoustic properties rather than any single dominant factor. For example, variation in vocal shimmer (perceived as a change in vocal qualities, such as breathiness~\cite{shimmer}) has a positive effect on appeal, whereas higher spectral flux (correlated to speaker gender~\cite{spectral_flux_gender}) is less favourable. Modelling narrator characteristics (e.g., narrator gender) instead of the acoustic correlates could clarify the driving effects and improve interpretability. We find that the duration sum correlates with higher view-rates; this is likely explained by longer audiobooks requiring more page visits to complete rather than a preference for longer recordings. Articulation rate (syllables\_per\_min\_std, which correlates with syllables\_per\_min\_mean: $\rho=0.66$) has a positive influence on appeal; this aligns with findings from a user study~\cite{narrative_lange_2020}, showing that higher articulation rates increase listener absorption and liking. Further analysis of feature correlations is required to refine interpretation. \newline
{\bf \noindent Genre-specific}:
The influence of acoustic features on appeal varies across genres, both in magnitude and direction. For example, variation in vocal shimmer (perceived as breathiness~\cite{shimmer}), which shows a positive effect globally ($\beta = 0.09$), is considerably stronger within Romance ($\beta = 0.31$), suggesting that this vocal characteristic is particularly relevant to appeal in this genre. In contrast, vocal shimmer shows no significant effect on appeal within History. Instead, variation in the Hammarberg index, a spectral measure of vocal quality associated with vocal effort and emotion~\cite{Ekberg2025Acoustic}, has the strongest effect on appeal within History ($\beta = -0.35$), compared to a weak global effect ($\beta = -0.04$) and no significant effect on Romance. Further analysis of outliers is required before drawing firm conclusions about their influence on appeal (see SM:Fig. 1). \newline
{\bf \noindent Intra-title}:
Variation in appeal across narrations of the same title (0.52) is nearly as large as variation across different titles (0.54), indicating that narration contributes on a scale comparable to content-level differences. This underscores the importance of vocal delivery in influencing listener engagement. This holds even if we limit the data to audiobooks with more than one version. Furthermore, the mixed-effects model substantially improves model fit relative to the global GLM ($AIC_{GLM} - AIC_{LME} = 210$). This highlights the importance of modelling title biases, which likely influence selection behaviour and page-view rates. Despite the better fit, most features that were statistically significant from the global GLM, remain significant after modelling effects driven by title (SM:Fig. 1). While the magnitude of the coefficients has a small amount of variation, the directions remain consistent.

\vspace{-6pt}
\subsection{Predictive Modelling Results}
\vspace{-4pt}

\begin{table}[!t]
    \centering
    \footnotesize
    \renewcommand{\arraystretch}{1.15}

    \caption{Accuracy [95\% CIs] of quartile classification models.}

    \begin{tabularx}{\linewidth}{@{} l c c c @{}}
        \toprule
        & \textbf{Genres} & \textbf{Audio Features} & \textbf{Combined} \\
        \midrule
        LR
        & 0.31 [0.31, 0.32]
        & \textbf{0.32 [0.32, 0.33]}
        & \textbf{0.35 [0.34, 0.35]} \\

        MLP
        & 0.31 [0.31, 0.32]
        & 0.30 [0.29, 0.30]
        & 0.32 [0.31, 0.33] \\

        SVM
        & \textbf{0.32 [0.31, 0.33]}
        & 0.31 [0.30, 0.32]
        & 0.33 [0.32, 0.34] \\

        XGBoost
        & \textbf{0.32 [0.31, 0.33]}
        & 0.29 [0.28, 0.30]
        & 0.31 [0.31, 0.32] \\

        Random
        & 0.25 [0.24, 0.26]
        & 0.25 [0.24, 0.26]
        & 0.25 [0.24, 0.26] \\
        \bottomrule
    \end{tabularx}

    \label{tab:classification}
\end{table}

\begin{table}[!t]
    \caption{Kendall's $\tau$ [95\% CIs] for ranking models. VR = view-rate, RR = return-rate.}
    \centering
    \footnotesize
    \renewcommand{\arraystretch}{1}
    \setlength{\tabcolsep}{3.5pt}
    \resizebox{\columnwidth}{!}{%
    \begin{tabularx}{\linewidth}{@{} l l l l @{}}
        \toprule
        & Full (VR) 
        & Subset (VR) 
        & Subset (RR) \\
        \midrule
        NDCG     & 0.08 [-0.03, 0.17] & -0.02 [-0.18, 0.13] & \textbf{0.26} [0.11, 0.41] \\
        Pair     & 0.10 [0.00, 0.20] & 0.01 [-0.14, 0.16] & \textbf{0.26} [0.11, 0.42] \\
        Lambda  & 0.13 [0.03, 0.23] & 0.02 [-0.13, 0.17] & \textbf{0.28} [0.13, 0.42] \\
        LR (group)   & 0.09 [0.02, 0.15] & 0.02 [-0.08, 0.12] & 0.08 [-0.01, 0.18] \\
        LR (full) & 0.10 [0.03, 0.17] & 0.04 [-0.04, 0.13] & 0.07 [-0.02, 0.17] \\
        Random         & 0.00 [-0.15, 0.14] & -- & -- \\
        \bottomrule
    \end{tabularx}
    } 
    \vspace{0.4em}
    \raggedright
    \footnotesize   
    \label{tab:ranking}
\end{table}

{\bf \noindent Classification}:
All models performed above the random baseline of 0.25 (see Table~\ref{tab:classification}). Predicting appeal using acoustic features alone can improve accuracy by up to 0.07, reinforcing the findings from our GLM experiments: in spite of the challenging appeal data, acoustic features have a robust predictive power. Genre-only and audio-only models achieved similar performance, while combining them boosts performance up to 0.35, indicating that acoustic features and genres capture related but not identical information. Class-specific analyses further reveal that acoustic features capture nuances in the middle quartiles, whereas genre features contributed more strongly to predictions at the extremes (see SM:Fig. 2). Simpler models such as LR and SVM generally outperform more complex approaches (MLP, XGBoost). \newline
{\bf \noindent Ranking}:
All ranking models show a statistically significant improvement over a random baseline (Table~\ref{tab:ranking}), consistent with the other experiments, confirming an influence of acoustic features on appeal. We find that the pointwise models (LR) tend to perform similarly to other ranking models, even when trained on the full dataset (which has 10x the amount of samples to learn from). The ranking models benefit from modelling relative appeal, as it removes any title-specific biases in preferences. We note that the performance of the ranking models are sensitive to data shuffling across the folds; further analysis and larger, more granular data on engagement and appeal will help understand which ranking model is more appropriate for the task.

\vspace{-6pt}
\subsection{Analysis Using Proprietary Engagement Metrics}
\label{sec:proprietary_experiment}
\vspace{-2pt}

While LibriVox view-rate is publicly available, which ensures that the results in this work are reproducible by all, it is extremely coarse and has many limitations, as discussed above. Thus, we recompute our analysis on a subset of LibriVox audiobooks (Sec.~\ref{sec:librivox}) that are also hosted on Spotify, for which we have access to granular consumption data. We replace view-rate with the proportion of distinct users who return to an audiobook within 14 days (i.e. \textit{return-rate}). This metric indicates whether a listener enjoyed an audiobook enough to return and continue listening. Return-rate is also biased against shorter recordings, which may be completed in a single session; developing more nuanced engagement metrics remains an important direction for future work. This data is available for 3{,}428 audiobooks, a subset of the original dataset (8{,}854 items).

 \noindent\textbf{Global GLM:} Refitting the GLM on this subset using view-rate increases pseudo-$R^{2}$ from 0.09 to 0.13, reflecting differences in sample composition. Replacing view-rate with the number of returning users within 14 days and setting total users as an exposure offset further improves performance (pseudo-$R^{2}=0.16$) and substantially improves model fit ($\Delta AIC \approx 6000$). These results suggest that audio features are more strongly associated with engagement conditional on exposure than raw popularity.
 
 \noindent\textbf{Ranking:} The LibriVox ranking experiment includes 736 audiobooks across 305 book groups (Sec.~\ref{sec:pred_modelling}:Ranking); this subset has 327 audiobooks in 138 groups. In this reduced setting, using view-rate fails to learn a meaningful relationship with audio features (Kendall's $\tau \approx 0$). In contrast, using return-rate to define relative appeal yields a strong and consistent relationship ($\tau \approx 0.26$--$0.28$), indicating that return-rate captures a more stable and discriminative intra-title appeal (see Table~\ref{tab:ranking}).

\section{Conclusion}
\label{sec:conclusion}

We examined the relationship between audiobook narration, genres, title, and consumption, and consistently found that acoustic features of narration influence appeal. The robustness of these results, despite coarse consumption data and mixed recording quality, validates our hypothesis that narration styles influence appeal, and point the way to exciting further research. 

Modelling relative appeal within book-groups emphasises the substantial contribution of narrations beyond content-level differences. We also find that the influence of different acoustic features varies across genres. Both findings highlight the need to account for title and genre information alongside audio features. While specific acoustic effects should be interpreted cautiously given the dataset limitations and feature correlations, several interesting interpretable patterns emerge. For example, greater variation in articulation rate is positively associated with appeal, while higher spectral flux (linked to vocal characteristics such as perceived gender) tends to be negatively associated. These trends show that measurable vocal and acoustic properties contribute to listener engagement beyond content-level factors. Exploratory analyses using more granular proprietary engagement data further support the conclusion that narration influences appeal, suggesting that this effect may be more pronounced using richer behavioural data. Together, these results provide converging evidence linking paralinguistic and acoustic features of narration to audiobook appeal and establish a foundation for future work using larger, more diverse datasets. 

Future work with richer engagement data (e.g., completion behaviour, user journeys across narrations and titles) and broader catalogues including professional narrations will enable a deeper understanding of narration appeal. Testing expanded feature sets, such as perceived narrator characteristics (accent, age, gender), whether the author narrates their own work, and how vocal performance and characterisation suit the narrative tone and context, will support more holistic and interpretable models. Analysing appeal across demographic listener segments will allow us to uncover audience-specific patterns of narration preference, which we hypothesise also shape engagement. Such analyses will allow us to move beyond coarse correlates and towards nuanced models that can be employed to improve personalisation, promotions, casting and other downstream uses.

\clearpage

\section{Acknowledgments}

\ifcameraready
     We thank R. Dall, R. Jones, D. Korkinof, A. Lima, A. McDowell, S. Reddy, B. Regan, A. Torrisi, L. Vongsathorn, J. Walker, H. Zhang, E. zu Erbach for their useful feedback.
\else
     Anonymous acknowledgments.
\fi

\section{Generative AI Use Disclosure}
Generative AI tools were used to assist with language editing, formatting, and improving clarity of the manuscript. All experimental design, analysis, and results were conducted and verified by the authors.

\bibliographystyle{IEEEtran}
\bibliography{refs}

@inproceedings{sini-etal-2022-investigating,
    title = "Investigating Inter- and Intra-speaker Voice Conversion using Audiobooks",
    author = "Sini, Aghilas  and
      Lolive, Damien  and
      Barbot, Nelly  and
      Alain, Pierre",
    booktitle = "Proc. of the 13th LREC",
    month = jun,
    year = "2022",
    address = "Marseille, France",
    publisher = "European Language Resources Association",
    url = "https://aclanthology.org/2022.lrec-1.794/",
    pages = "7305--7313"
}

@mastersthesis{embretsen2019representing,
  title     = {Representing voices using convolutional neural network embeddings},
  author    = {Embrets{\'{e}}n, Niklas},
  year      = {2019},
  school    = {KTH, School of Electrical Engineering and Computer Science (EECS)},
  type      = {Master's thesis}
}

@article{audiobooks_role_of_narrator,
  title={Why Do We Listen to Audiobooks? The Role of Narrator Performance, BGM, Telepresence, and Emotional Connectedness},
  author={Ji, Dan and Liu, Boquan and Xu, Jinghong and Gong, Jiankun},
  journal={Sage Open},
  volume={14},
  number={2},
  year={2024},
  publisher={SAGE Publications Sage CA: Los Angeles, CA}
}

@article{yamnet,
  title={Mobilenets: Efficient convolutional neural networks for mobile vision applications},
  author={Howard, Andrew G and Zhu, Menglong and Chen, Bo and Kalenichenko, Dmitry and Wang, Weijun and Weyand, Tobias and Andreetto, Marco and Adam, Hartwig},
  journal={arXiv preprint arXiv:1704.04861},
  year={2017}
}

@article{JALALINAJAFABADI2021103018,
title = {Acoustic analysis and digital signal processing for the assessment of voice quality},
journal = {Biomedical Signal Processing and Control},
volume = {70},
pages = {103018},
year = {2021},
issn = {1746-8094},
doi = {https://doi.org/10.1016/j.bspc.2021.103018},
url = {https://www.sciencedirect.com/science/article/pii/S1746809421006157},
author = {Farideh Jalali-najafabadi and Chaitanya Gadepalli and Delaram Jarchi and Barry M.G. Cheetham}
}

@inproceedings{opensmile,
author = {Eyben, Florian and W\"{o}llmer, Martin and Schuller, Bj\"{o}rn},
title = {Opensmile: the munich versatile and fast open-source audio feature extractor},
year = {2010},
isbn = {9781605589336},
publisher = {Association for Computing Machinery},
address = {New York, NY, USA},
url = {https://doi.org/10.1145/1873951.1874246},
doi = {10.1145/1873951.1874246},
booktitle = {Proceedings of the 18th ACM International Conference on Multimedia},
pages = {1459–1462},
numpages = {4},
location = {Firenze, Italy},
series = {MM '10}
}

@ARTICLE{opensmile_gemaps,
  author={Eyben, Florian and Scherer, Klaus R. and Schuller, Bj{\"o}rn W. and Sundberg, Johan and Andr{\'e}, Elisabeth and Busso, Carlos and Devillers, Laurence Y. and Epps, Julien and Laukka, Petri and Narayanan, Shrikanth S. and Truong, Khiet P.},
  journal={IEEE Transactions on Affective Computing}, 
  title={The {Geneva} Minimalistic Acoustic Parameter Set {(GeMAPS)} for Voice Research and Affective Computing}, 
  year={2016},
  volume={7},
  number={2},
  pages={190-202},
  doi={10.1109/TAFFC.2015.2457417}}

@article{experiencing_kosch_2024,
    url = {https://doi.org/10.1515/jlt-2024-2005},
    title = {Experiencing Literary Audiobooks: A Framework for Theoretical and Empirical Investigations of the Auditory Reception of Literature},
    author = {Lukas Kosch and Annika Schwabe and Hajo Boomgaarden and G{\"u}nther Stocker},
    pages = {67--88},
    volume = {18},
    number = {1},
    journal = {Journal of Literary Theory},
    doi = {doi:10.1515/jlt-2024-2005},
    year = {2024},
}

@mastersthesis{Dakic1396778,
   author = {Dakic, Martina},
   school = {University of Bor{\aa}s, Faculty of Librarianship, Information, Education and IT},
   title = {{Preferences and attitudes of audiobook users in Sweden : Surveying Swedish audiobook groups on Facebook}},
   year = {2019}
}

@article{clustering_szkely_2011,
	title = {Clustering Expressive Speech Styles in Audiobooks Using Glottal Source Parameters.},
	doi = {10.21437/interspeech.2011-627},
	author = {Sz{\'e}kely, {\'E}va and Cabral, Jo{\~a}o P. and Cahill, Peter and Carson-Berndsen, Julie},
	year = {2011},
    journal = {Proc. Interspeech 2011},
    pages = {2409-2412}
}

@article{role_montao_2016,
title = {The role of prosody and voice quality in indirect storytelling speech: Annotation methodology and expressive categories},
journal = {Speech Communication},
volume = {85},
pages = {8-18},
year = {2016},
issn = {0167-6393},
doi = {https://doi.org/10.1016/j.specom.2016.10.006},
url = {https://www.sciencedirect.com/science/article/pii/S0167639315300108},
author = {Ra{\'u}l Monta{\~n}o and Francesc Al{\'i}as}
}

@article{role_montao_2017,
    title = {The role of prosody and voice quality in indirect storytelling speech: A cross-narrator perspective in four {European} languages},
    journal = {Speech Communication},
    volume = {88},
    pages = {1-16},
    year = {2017},
    issn = {0167-6393},
    doi = {https://doi.org/10.1016/j.specom.2017.01.007},
    url = {https://www.sciencedirect.com/science/article/pii/S0167639315300418},
    author = {Ra{\'u}l Monta{\~n}o and Francesc Al{\'i}as}
}

@INPROCEEDINGS{prosody_pethe_2023,
  author={Pethe, Charuta and Pham, Bach and Childress, Felix D and Yin, Yunting and Skiena, Steven},
  booktitle={2025 19th International Conference on Semantic Computing (ICSC)}, 
  title={Prosody Analysis of Audiobooks}, 
  year={2025},
  volume={},
  number={},
  pages={217-221},
  doi={10.1109/ICSC64641.2025.00036}}

@Article{narrative_lange_2020,
    author={Lange, Elke B.
    and Thiele, Dominik
    and Kuijpers, Moniek M.},
    title={Narrative aesthetic absorption in audiobooks is predicted by blink rate and acoustic features.},
    journal={Psychology of Aesthetics, Creativity, and the Arts},
    year={2022},
    publisher={Educational Publishing Foundation},
    address={US},
    volume={16},
    number={1},
    pages={110-124},
    doi={10.1037/aca0000321},
    url={https://doi.org/10.1037/aca0000321}
}

@inproceedings{evaluating_szkely_2012,
    title = "Evaluating expressive speech synthesis from audiobook corpora for conversational phrases",
    author = "Sz{\'e}kely, {\'E}va  and
      Cabral, Jo{\~a}o Paulo  and
      Abou-Zleikha, Mohamed  and
      Cahill, Peter  and
      Carson-Berndsen, Julie",
    editor = "Calzolari, Nicoletta  and
      Choukri, Khalid  and
      Declerck, Thierry  and
      Do{\u{g}}an, Mehmet U{\u{g}}ur  and
      Maegaard, Bente  and
      Mariani, Joseph  and
      Moreno, Asuncion  and
      Odijk, Jan  and
      Piperidis, Stelios",
    booktitle = "Proceedings of the Eighth International Conference on Language Resources and Evaluation ({LREC}'12)",
    month = may,
    year = "2012",
    address = "Istanbul, Turkey",
    publisher = "European Language Resources Association (ELRA)",
    url = "https://aclanthology.org/L12-1513/",
    pages = "3335--3339"
}

@article{synthetic_rodero_2021,
    author = {Emma Rodero and Ignacio Lucas},
    title ={Synthetic versus human voices in audiobooks: The human emotional intimacy effect},
    journal = {New Media \& Society},
    volume = {25},
    number = {7},
    pages = {1746-1764},
    year = {2023},
    doi = {10.1177/14614448211024142},
    URL = {https://doi.org/10.1177/14614448211024142},
    eprint = {https://doi.org/10.1177/14614448211024142}
}

@incollection{neurocomputational_krger_2018,
    author = {Kr{\"o}ger, Bernd J.},
    editor = {Fr{\"u}hholz, Sascha and Belin, Pascal},
    isbn = {9780198743187},
    title = {Neurocomputational Models of Voice and Speech Perception},
    booktitle = {The Oxford Handbook of Voice Perception},
    publisher = {Oxford University Press},
    year = {2018},
    month = {12},
    doi = {10.1093/oxfordhb/9780198743187.013.34},
    url = {https://doi.org/10.1093/oxfordhb/9780198743187.013.34},
    eprint = {https://academic.oup.com/book/0/chapter/335932391/chapter-ag-pdf/44442005/book_38687_section_335932391.ag.pdf},
}

@Inbook{multidimensional_obuchi_2021,
    author="Obuchi, Yasunari",
    editor="Weiss, Benjamin
    and Trouvain, J{\"u}rgen
    and Barkat-Defradas, Melissa
    and Ohala, John J.",
    title="Multidimensional Mapping of Voice Attractiveness and Listener's Preference: Optimization and Estimation from Audio Signal",
    bookTitle="Voice Attractiveness: Studies on Sexy, Likable, and Charismatic Speakers",
    year="2021",
    publisher="Springer Singapore",
    address="Singapore",
    pages="281--295",
    isbn="978-981-15-6627-1",
    doi="10.1007/978-981-15-6627-1_15",
    url="https://doi.org/10.1007/978-981-15-6627-1_15"
}

@article{affective_weningerd_2022,
    title = {Affective and behavioural computing: Lessons learnt from the First Computational Paralinguistics Challenge},
    journal = {Computer Speech \& Language},
    volume = {53},
    pages = {156-180},
    year = {2019},
    issn = {0885-2308},
    doi = {https://doi.org/10.1016/j.csl.2018.02.004},
    url = {https://www.sciencedirect.com/science/article/pii/S0885230816303928},
    author = {Bj{\"o}rn Schuller and Felix Weninger and Yue Zhang and Fabien Ringeval and Anton Batliner and Stefan Steidl and Florian Eyben and Erik Marchi and Alessandro Vinciarelli and Klaus Scherer and Mohamed Chetouani and Marcello Mortillaro}
}

@INPROCEEDINGS{scripted_spontaneous_elisha_2024,
  author={Elisha, Shahar and McDowell, Andrew and Beguerisse-D{\'i}az, Mariano and Benetos, Emmanouil},
  booktitle={2024 SLT}, 
  title={Classification Of Spontaneous And Scripted Speech For Multilingual Audio}, 
  year={2024},
  volume={},
  number={},
  pages={489-495},
  doi={10.1109/SLT61566.2024.10832309}
}

@INPROCEEDINGS{ser_generalisation_goron_2024,
  author={Goron, Erik and Asai, Lena and Rut, Elias and Dinov, Martin},
  booktitle={ICASSP 2024}, 
  title={Improving Domain Generalization in Speech Emotion Recognition with {Whisper}}, 
  year={2024},
  volume={},
  number={},
  pages={11631-11635},
  doi={10.1109/ICASSP48485.2024.10446997}
}

@article{ethical_awareness_batliner_2023,
    author = {Anton Batliner and Michael Neumann and Felix Burkhardt and Alice Baird and Sarina Meyer and Ngoc Thang Vu and Bj{\"o}rn W. Schuller},
    title = {Ethical Awareness in Paralinguistics: A Taxonomy of Applications},
    journal = {International Journal of Human–Computer Interaction},
    volume = {39},
    number = {9},
    pages = {1904--1921},
    year = {2023},
    publisher = {Taylor \& Francis},
    doi = {10.1080/10447318.2022.2140385},
    URL = {https://doi.org/10.1080/10447318.2022.2140385}
}

@inproceedings{whisper,
author = {Radford, Alec and Kim, Jong Wook and Xu, Tao and Brockman, Greg and McLeavey, Christine and Sutskever, Ilya},
title = {Robust speech recognition via large-scale weak supervision},
year = {2023},
publisher = {JMLR.org},
booktitle = {Proceedings of the 40th International Conference on Machine Learning},
articleno = {1182},
numpages = {27},
location = {Honolulu, Hawaii, USA},
series = {ICML'23}
}

@INPROCEEDINGS{audioset,
  author={Gemmeke, Jort F. and Ellis, Daniel P. W. and Freedman, Dylan and Jansen, Aren and Lawrence, Wade and Moore, R. Channing and Plakal, Manoj and Ritter, Marvin},
  booktitle={2017 IEEE International Conference on Acoustics, Speech and Signal Processing (ICASSP)}, 
  title={Audio Set: An ontology and human-labeled dataset for audio events}, 
  year={2017},
  volume={},
  number={},
  pages={776-780},
  doi={10.1109/ICASSP.2017.7952261}}

@techreport{burges2010from,
author = {Burges, Chris J.C.},
title = {From {RankNet to LambdaRank to LambdaMART}: An Overview},
year = {2010},
month = {June},
url = {https://www.microsoft.com/en-us/research/publication/from-ranknet-to-lambdarank-to-lambdamart-an-overview/},
number = {MSR-TR-2010-82},
}

@INPROCEEDINGS{spectral_flux_gender,
  author={Yasmin, Ghazaala and Dutta, Suchibrota and Ghosal, Arijit},
  booktitle={2017 International Conference on Intelligent Computing, Instrumentation and Control Technologies (ICICICT)}, 
  title={Discrimination of male and female voice using occurrence pattern of spectral flux}, 
  year={2017},
  volume={},
  number={},
  pages={576-581},
  doi={10.1109/ICICICT1.2017.8342627}}

@INPROCEEDINGS{audiobook_tts,
  author={Manoj, Balu and Jiji, Justin and Dileep, Rahul and Manohar, Nandana},
  booktitle={2025 International Conference on Computing Technologies (ICOCT)}, 
  title={Emotionally Enhanced Audiobook Reader with Character Voice Differentiation}, 
  year={2025},
  volume={},
  number={},
  pages={1-6},
  doi={10.1109/ICOCT64433.2025.11118389}}

@INPROCEEDINGS{charisma,
  author={Kathan, Alexander and Amiriparian, Shahin and Christ, Lukas and Eulitz, Simone and Schuller, Bj{\"o}rn W.},
  booktitle={2024 IEEE Conference on Telepresence}, 
  title={Automatic Speech-Based Charisma Recognition and the Impact of Integrating Auxiliary Characteristics}, 
  year={2024},
  volume={},
  number={},
  pages={148-153},
  doi={10.1109/Telepresence63209.2024.10841640}}

@ARTICLE{depression,
  author={Leal, Samara Soares and Ntalampiras, Stavros and Sassi, Roberto},
  journal={IEEE Transactions on Affective Computing}, 
  title={Speech-Based Depression Assessment: A Comprehensive Survey}, 
  year={2025},
  volume={16},
  number={3},
  pages={1318-1333},
  doi={10.1109/TAFFC.2024.3521327}}

@book{schuller2013_computational_paralinguistics,
  title={Computational paralinguistics: emotion, affect and personality in speech and language processing},
  author={Schuller, Bj{\"o}rn and Batliner, Anton},
  year={2013},
  publisher={John Wiley \& Sons}
}

@inproceedings{diffkendall,
author = {Zheng, Kaipeng and Zhang, Huishuai and Huang, Weiran},
title = {{DiffKendall}: a novel approach for few-shot learning with differentiable kendall's rank correlation},
year = {2023},
publisher = {Curran Associates Inc.},
address = {Red Hook, NY, USA},
booktitle = {Proc. of the 37th NeurIPS},
articleno = {2149},
numpages = {13},
location = {New Orleans, LA, USA},
series = {NIPS '23}
}

@book{montgomery2021introduction,
  title={Introduction to Linear Regression Analysis},
  author={Montgomery, Douglas C and Peck, Elizabeth A and Vining, G. Geoffrey},
  year={2021},
  publisher={Wiley},
  edition={6}
}

@article{benjamini-hochberg,
author = {Benjamini, Yoav and Hochberg, Yosef},
title = {Controlling the False Discovery Rate: A Practical and Powerful Approach to Multiple Testing},
journal = {Journal of the Royal Statistical Society: Series B (Methodological)},
volume = {57},
number = {1},
pages = {289-300},
doi = {https://doi.org/10.1111/j.2517-6161.1995.tb02031.x},
url = {https://rss.onlinelibrary.wiley.com/doi/abs/10.1111/j.2517-6161.1995.tb02031.x},
eprint = {https://rss.onlinelibrary.wiley.com/doi/pdf/10.1111/j.2517-6161.1995.tb02031.x},
year = {1995}
}

@ARTICLE{aic,
  author={Akaike, H.},
  journal={IEEE Transactions on Automatic Control}, 
  title={A new look at the statistical model identification}, 
  year={1974},
  volume={19},
  number={6},
  pages={716-723},
  doi={10.1109/TAC.1974.1100705}}

@article{netflix_recs,
author = {Gomez-Uribe, Carlos A. and Hunt, Neil},
title = {The {Netflix} Recommender System: Algorithms, Business Value, and Innovation},
year = {2016},
issue_date = {January 2016},
publisher = {Association for Computing Machinery},
address = {New York, NY, USA},
volume = {6},
number = {4},
issn = {2158-656X},
url = {https://doi.org/10.1145/2843948},
doi = {10.1145/2843948},
journal = {ACM Trans. Manage. Inf. Syst.},
month = dec,
articleno = {13},
numpages = {19}
}

@inproceedings{lambdamart,
author = {Burges, Christopher J. C. and Ragno, Robert and Le, Quoc Viet},
title = {Learning to rank with nonsmooth cost functions},
year = {2006},
publisher = {MIT Press},
address = {Cambridge, MA, USA},
booktitle = {Proceedings of the 20th International Conference on Neural Information Processing Systems},
pages = {193–200},
numpages = {8},
location = {Canada},
series = {NIPS'06}
}

@inproceedings{ranknet,
author = {Burges, Chris and Shaked, Tal and Renshaw, Erin and Lazier, Ari and Deeds, Matt and Hamilton, Nicole and Hullender, Greg},
title = {Learning to rank using gradient descent},
year = {2005},
isbn = {1595931805},
publisher = {Association for Computing Machinery},
address = {New York, NY, USA},
url = {https://doi.org/10.1145/1102351.1102363},
doi = {10.1145/1102351.1102363},
booktitle = {Proceedings of the 22nd International Conference on Machine Learning},
pages = {89–96},
numpages = {8},
location = {Bonn, Germany},
series = {ICML '05}
}

@article{Ekberg2025Acoustic,
  author  = {Ekberg, M. and Stavrinos, G. and Andin, J. and Stenfelt, S. and Dahlstr{\"o}m, {\"O}.},
  title   = {Acoustic Features Distinguishing Emotions in {Swedish} Speech},
  journal = {Journal of Voice},
  year    = {2025},
  volume  = {39},
  number  = {6},
  pages   = {1699.e11--1699.e20},
  doi     = {10.1016/j.jvoice.2023.03.010},
  publisher = {Elsevier},
  issn    = {0892-1997},
  url     = {https://doi.org/10.1016/j.jvoice.2023.03.010}
}

@misc{librivox,
  author       = {{LibriVox}},
  title        = {{LibriVox}: Free Public Domain Audiobooks},
  year         = {2025},
  howpublished = {\url{https://librivox.org}},
}

@misc{internetarchive_librivox,
  author       = {{Internet Archive}},
  title        = {{LibriVox} Audio Collection},
  year         = {2025},
  howpublished = {\url{https://archive.org/details/librivoxaudio}},
}

@misc{syllables_package,
  author       = {Kyle Gorman},
  title        = {syllables: A Simple Syllable Counting Package for {Python}},
  year         = {2025},
  howpublished = {\url{https://pypi.org/project/syllables/}},
}

@misc{xgboost_docs,
  author       = {{XGBoost Developers}},
  title        = {{XGBoost} Documentation},
  year         = {2025},
  howpublished = {\url{https://xgboost.readthedocs.io}},
}

@misc{lightgbm_docs,
  author       = {{LightGBM Developers}},
  title        = {{LightGBM} Documentation},
  year         = {2025},
  howpublished = {\url{https://lightgbm.readthedocs.io}},
}

@inproceedings{spotify_recs,
author = {Fazelnia, Ghazal and Gupta, Sanket and Keum, Claire and Koh, Mark and Heath, Timothy and Carrasco Hern\'{a}ndez, Guillermo and Xie, Stephen and Singh, Nandini and Anderson, Ian and Hristakeva, Maya and Pehrson Skid\'{e}n, Petter and Lalmas, Mounia},
title = {Generalized User Representations for Large-Scale Recommendations and Downstream Tasks},
year = {2025},
isbn = {9798400713644},
publisher = {Association for Computing Machinery},
address = {New York, NY, USA},
url = {https://doi.org/10.1145/3705328.3748132},
doi = {10.1145/3705328.3748132},
booktitle = {Proceedings of the Nineteenth ACM Conference on Recommender Systems},
pages = {962–966},
numpages = {5},
location = {
},
series = {RecSys '25}
}

@article{shimmer,
title = {The Effect of Levels and Types of Experience on Judgment of Synthesized Voice Quality},
journal = {Journal of Voice},
volume = {28},
number = {1},
pages = {24-35},
year = {2014},
issn = {0892-1997},
doi = {https://doi.org/10.1016/j.jvoice.2013.06.001},
url = {https://www.jvoice.org/article/S0892-1997(13)00103-3/abstract},
author = {Jessica L. Sofranko and Robert A. Prosek},
}

\end{document}


\onecolumn
\maketitle

\begin{table*}[h]
    \centering
    \caption{Complete feature set used in the experiments (129 features), grouped by extraction method.}
    
    \begin{tabular}{ll}
    \toprule
    \multicolumn{2}{l}{\textbf{eGeMAPSv02}}~\cite{opensmile_gemaps} \\ 
    \midrule
    
    F0semitoneFrom27.5Hz\_sma3nzV\_amean & F1amplitudeLogRelF0\_sma3nzV\_amean \\
    F0semitoneFrom27.5Hz\_sma3nzV\_stddevNorm & F1amplitudeLogRelF0\_sma3nzV\_stddevNorm \\
    F0semitoneFrom27.5Hz\_sma3nz\_percentile20.0 & F2frequency\_sma3nzV\_amean \\
    F0semitoneFrom27.5Hz\_sma3nz\_percentile50.0 & F2frequency\_sma3nzV\_stddevNorm \\
    F0semitoneFrom27.5Hz\_sma3nz\_percentile80.0 & F2bandwidth\_sma3nzV\_amean \\
    F0semitoneFrom27.5Hz\_sma3nz\_pctlrange0-2 & F2bandwidth\_sma3nzV\_stddevNorm \\
    F0semitoneFrom27.5Hz\_sma3nz\_meanRisingSlope & F2amplitudeLogRelF0\_sma3nzV\_amean \\
    F0semitoneFrom27.5Hz\_sma3nz\_stddevRisingSlope & F2amplitudeLogRelF0\_sma3nzV\_stddevNorm \\
    F0semitoneFrom27.5Hz\_sma3nz\_meanFallingSlope & F3frequency\_sma3nzV\_amean \\
    F0semitoneFrom27.5Hz\_sma3nz\_stddevFallingSlope & F3frequency\_sma3nzV\_stddevNorm \\
    
    F1frequency\_sma3nzV\_amean & F3bandwidth\_sma3nzV\_amean \\
    F1frequency\_sma3nzV\_stddevNorm & F3bandwidth\_sma3nzV\_stddevNorm \\
    F1bandwidth\_sma3nzV\_amean & F3amplitudeLogRelF0\_sma3nzV\_amean \\
    F1bandwidth\_sma3nzV\_stddevNorm & F3amplitudeLogRelF0\_sma3nzV\_stddevNorm \\
    
    Loudness\_sma3\_amean & alphaRatio\_sma3V\_amean \\
    Loudness\_sma3\_stddevNorm & alphaRatio\_sma3V\_stddevNorm \\
    Loudness\_sma3\_meanRisingSlope & alphaRatio\_sma3UV\_amean \\
    Loudness\_sma3\_stddevRisingSlope & hammarbergIndex\_sma3V\_amean \\
    Loudness\_sma3\_meanFallingSlope & hammarbergIndex\_sma3V\_stddevNorm \\
    Loudness\_sma3\_stddevFallingSlope & hammarbergIndex\_sma3UV\_amean \\
    
    spectralFlux\_sma3\_amean & spectralFlux\_sma3V\_amean \\
    spectralFlux\_sma3\_stddevNorm & spectralFlux\_sma3V\_stddevNorm \\
    spectralFlux\_sma3UV\_amean & slope0-500\_sma3V\_amean \\
    slope0-500\_sma3V\_stddevNorm & slope0-500\_sma3UV\_amean \\
    slope500-1500\_sma3V\_amean & slope500-1500\_sma3V\_stddevNorm \\
    slope500-1500\_sma3UV\_amean & equivalentSoundLevel\_dBp \\
    
    mfcc1\_sma3\_amean & mfcc1\_sma3V\_amean \\
    mfcc1\_sma3\_stddevNorm & mfcc1\_sma3V\_stddevNorm \\
    mfcc2\_sma3\_amean & mfcc2\_sma3V\_amean \\
    mfcc2\_sma3\_stddevNorm & mfcc2\_sma3V\_stddevNorm \\
    mfcc3\_sma3\_amean & mfcc3\_sma3V\_amean \\
    mfcc3\_sma3\_stddevNorm & mfcc3\_sma3V\_stddevNorm \\
    mfcc4\_sma3\_amean & mfcc4\_sma3V\_amean \\
    mfcc4\_sma3\_stddevNorm & mfcc4\_sma3V\_stddevNorm \\
    
    jitterLocal\_sma3nzV\_amean & shimmerLocaldB\_sma3nzV\_amean \\
    jitterLocal\_sma3nzV\_stddevNorm & shimmerLocaldB\_sma3nzV\_stddevNorm \\
    HNRdBACF\_sma3nzV\_amean & HNRdBACF\_sma3nzV\_stddevNorm \\
    logRelF0-H1-H2\_sma3nzV\_amean & logRelF0-H1-A3\_sma3nzV\_amean \\
    logRelF0-H1-H2\_sma3nzV\_stddevNorm & logRelF0-H1-A3\_sma3nzV\_stddevNorm \\
    
    VoicedSegmentsPerSec & MeanVoicedSegmentLengthSec \\
    StddevVoicedSegmentLengthSec & MeanUnvoicedSegmentLength \\
    StddevUnvoicedSegmentLength & LoudnessPeaksPerSecond \\
    
    \toprule
    \multicolumn{2}{l}{\textbf{Whisper-tiny}}~\cite{whisper} \\
    \midrule
    words\_per\_minute\_mean & syllables\_per\_minute\_mean \\
    words\_per\_minute\_std & syllables\_per\_minute\_std \\
    words\_per\_minute\_min & syllables\_per\_minute\_min \\
    words\_per\_minute\_max & syllables\_per\_minute\_max \\
    word\_count\_sum & syllable\_count\_sum \\
    duration\_sum & \\
    
    \toprule
    \multicolumn{2}{l}{\textbf{YAMNet}}~\cite{yamnet} \\ 
    \midrule
    
    speech\_mean & speech\_std \\
    child\_speech\_mean & child\_speech\_std \\
    conversation\_mean & conversation\_std \\
    narration\_monologue\_mean & narration\_monologue\_std \\
    babbling\_mean & babbling\_std \\
    speech\_synthesizer\_mean & speech\_synthesizer\_std \\
    shout\_mean & shout\_std \\
    bellow\_mean & bellow\_std \\
    whoop\_mean & whoop\_std \\
    yell\_mean & yell\_std \\
    children\_shouting\_mean & children\_shouting\_std \\
    screaming\_mean & screaming\_std \\
    whispering\_mean & whispering\_std \\
    music\_mean & music\_std \\
    sound\_effects\_mean & sound\_effects\_std \\
    non\_verbal\_vocalisations\_mean & non\_verbal\_vocalisations\_std \\
    recording\_quality\_mean & recording\_quality\_std \\
    \bottomrule
    \end{tabular}
    \label{tab:feature_set}
\end{table*}

\newpage

\begin{figure*}[htb]
  \centering
  \centerline{\includegraphics[width=\textwidth]{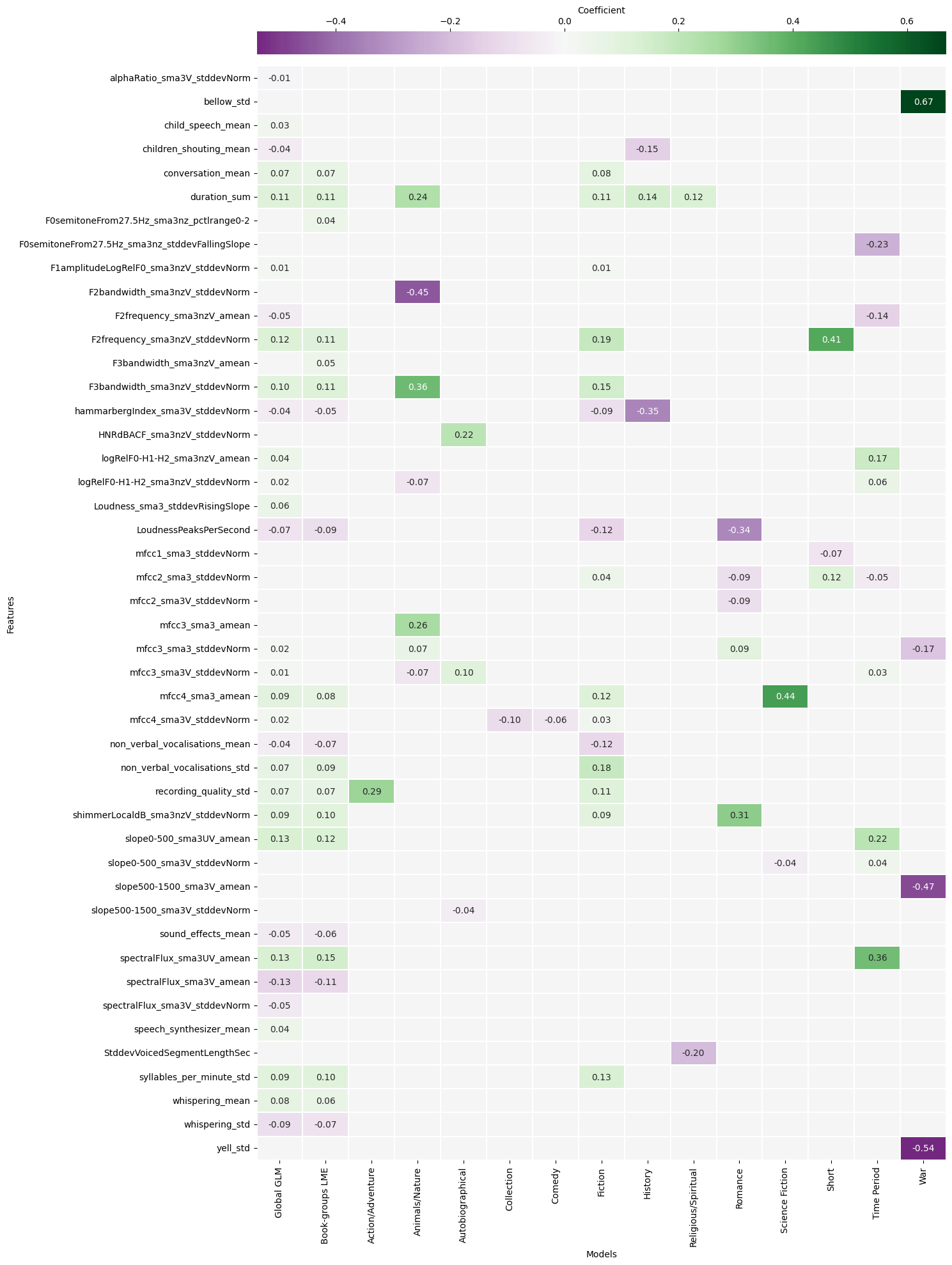}}
  \caption{Benjamini-Hochberg corrected, significant ($p < 0.05$) feature coefficients from the Global GLM, Book-groups LME, and genre GLMs. We exclude genres that have fewer than 350 audiobooks.}
  \label{fig:coefficients}
\end{figure*}

\begin{figure}[t]
\centering

\begin{subfigure}{0.32\linewidth}
    \centering
    \includegraphics[width=\linewidth]{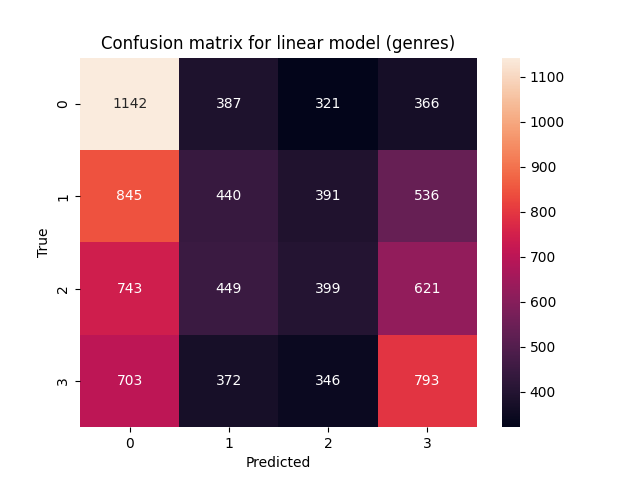}
    \caption{Genres}
\end{subfigure}
\hfill
\begin{subfigure}{0.32\linewidth}
    \centering
    \includegraphics[width=\linewidth]{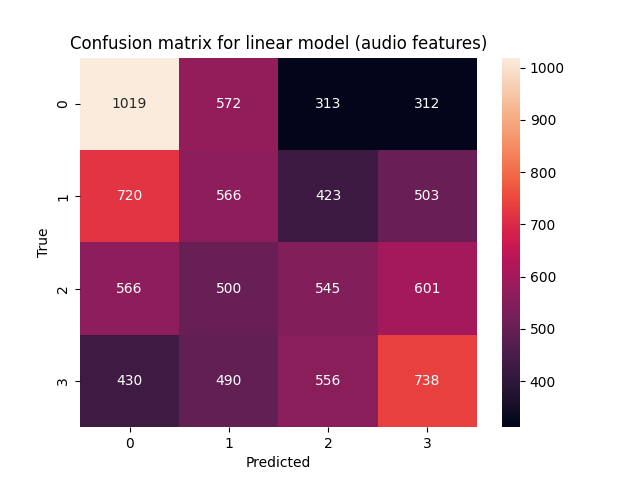}
    \caption{Audio-features}
\end{subfigure}
\hfill
\begin{subfigure}{0.32\linewidth}
    \centering
    \includegraphics[width=\linewidth]{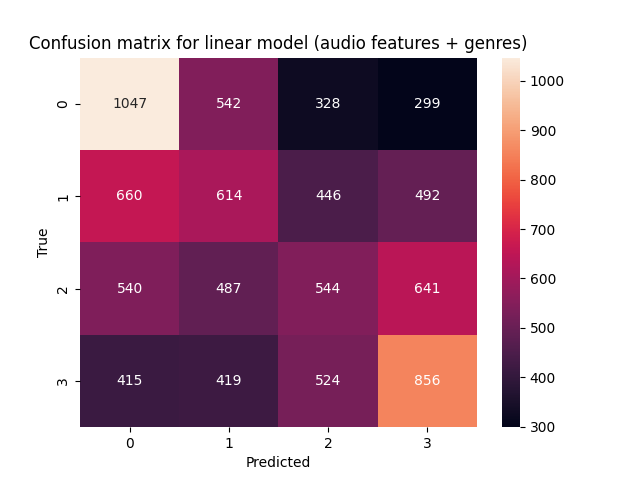}
    \caption{Audio-features and genres combined}
\end{subfigure}

\caption{Confusion matrices of the logisitic regression models trained on genres, audio-features, and a combination of both.}
\label{fig:heatmaps}
\end{figure}

\clearpage
\bibliographystyle{IEEEtran}
\bibliography{refs}